\documentclass[conference]{IEEEtran}
\usepackage{times}

\usepackage{microtype}
\usepackage{graphicx}
\usepackage{booktabs}
\usepackage{cite}
\usepackage{multirow}
\usepackage{tabularx}
\usepackage{amsmath}
\usepackage{amssymb}
\usepackage{mathtools}
\usepackage{amsthm}
\usepackage{comment}
\usepackage{subfigure}
\usepackage{pifont}
\usepackage{multicol}
\usepackage[bookmarks=true]{hyperref}
\newcommand{\framework}{{{\texttt{AID-SR}}}}

\begin{document}

\title{Bridging Language and Physics: Automated Design of Continuum Robots with Large Language Models}

\newcommand{\affilmark}[1]{\textsuperscript{#1}}
\newcommand{\equalcontrib}{\textsuperscript{\dag}}
\newcommand{\CA}{\textsuperscript{*}}

\author{
\authorblockN{
Jingyi Chen\affilmark{1}\equalcontrib,
Mohan Zhang\affilmark{2}\equalcontrib,
Laura Yao\affilmark{2}\equalcontrib, 
Yingtai Ni\affilmark{1},
Jianmin Ji\affilmark{1},
Jie Peng\affilmark{1},
Song Wang\affilmark{3},
and
Tianlong Chen\affilmark{2}\CA
}

\authorblockA{\affilmark{1}School of Computer Science and Technology, University of Science and Technology of China, Hefei, China}
\authorblockA{\affilmark{2}Department of Computer Science, University of North Carolina at Chapel Hill, Chapel Hill, USA}
\authorblockA{\affilmark{3}Department of Computer Science, University of Central Florida, Orlando, USA}

\authorblockA{
\textsuperscript{\dag}Equal contribution
\quad
\textsuperscript{*}Corresponding author
}
}

\maketitle

\begin{abstract}
Large language models (LLMs) have recently emerged as a promising tool for automating robot design from high-level specifications, yet they remain ineffective for robots operating under complex physical interactions. This limitation stems from the gap between language-based reasoning and the physical consequences of embodiment, often resulting in designs with low physical validity.
In this work, we propose a multi-layered framework, \framework, that establishes a closed loop by translating simulator-observed physical states into structured feedback for the LLM designer. Combined with semantic critique, human feedback, and iterative refinement, the framework promotes the generation of physically feasible and functionally meaningful robot designs. 
We evaluate our approach on tendon-driven continuum robots across a benchmark of 14 tasks spanning reaching, grasping, locomotion, and manipulation.
The proposed framework achieves $96.2\%$ rate for passing the simulation feasibility check and by applying a common reinforcement learning training, $26.7\%$ robots can successfully fulfill the corresponding task.
We then fabricate three designed robots of \framework\ that successfully complete the task in real-world.
These extensive experiments across simulation and real-world environments demonstrate and break the wall of utilizing the LLMs for automated design of continuum robots.
The source code and experimental resources are publicly available at \url{https://github.com/UNITES-Lab/AID-SR}.

\end{abstract}

\IEEEpeerreviewmaketitle

\section{Introduction}
The automated design of robotic systems has long been pursued as a way to reduce human effort and to accelerate the exploration of novel mechanical structures\cite{lipson2000automatic, alattas2019evolutionary, prabhu2018survey}. Traditionally, this process relies heavily on expert intuition and labor-intensive trial-and-error, particularly when robots operate under complex physical interactions such as large deformation, contact, and nonlinear dynamics\cite{rus2015design, hawkes2021hard}.

Recent progress in large language models (LLMs) has sparked growing interest in using language-driven generative models to automate robot design from high-level, natural language specifications. A number of recent studies have demonstrated that LLMs can generate syntactically correct simulation files, mechanical descriptions, or executable design code\cite{chen2025large, song2025laser, ma2024exploration, fang2025robomore, qiu2024robomorph}. While these studies demonstrate the potential of LLMs as design tools, most of them focuses on simplified settings or abstract tasks, such as 2-dimensional locomotion. As a result, the generated designs lack clear relevance to real-world manipulation or human-assistive scenarios, where robots must operate in 3-dimensional spaces and interact purposefully with external objects.

\begin{figure}[!t]
    \centering
    \includegraphics[width=\linewidth]{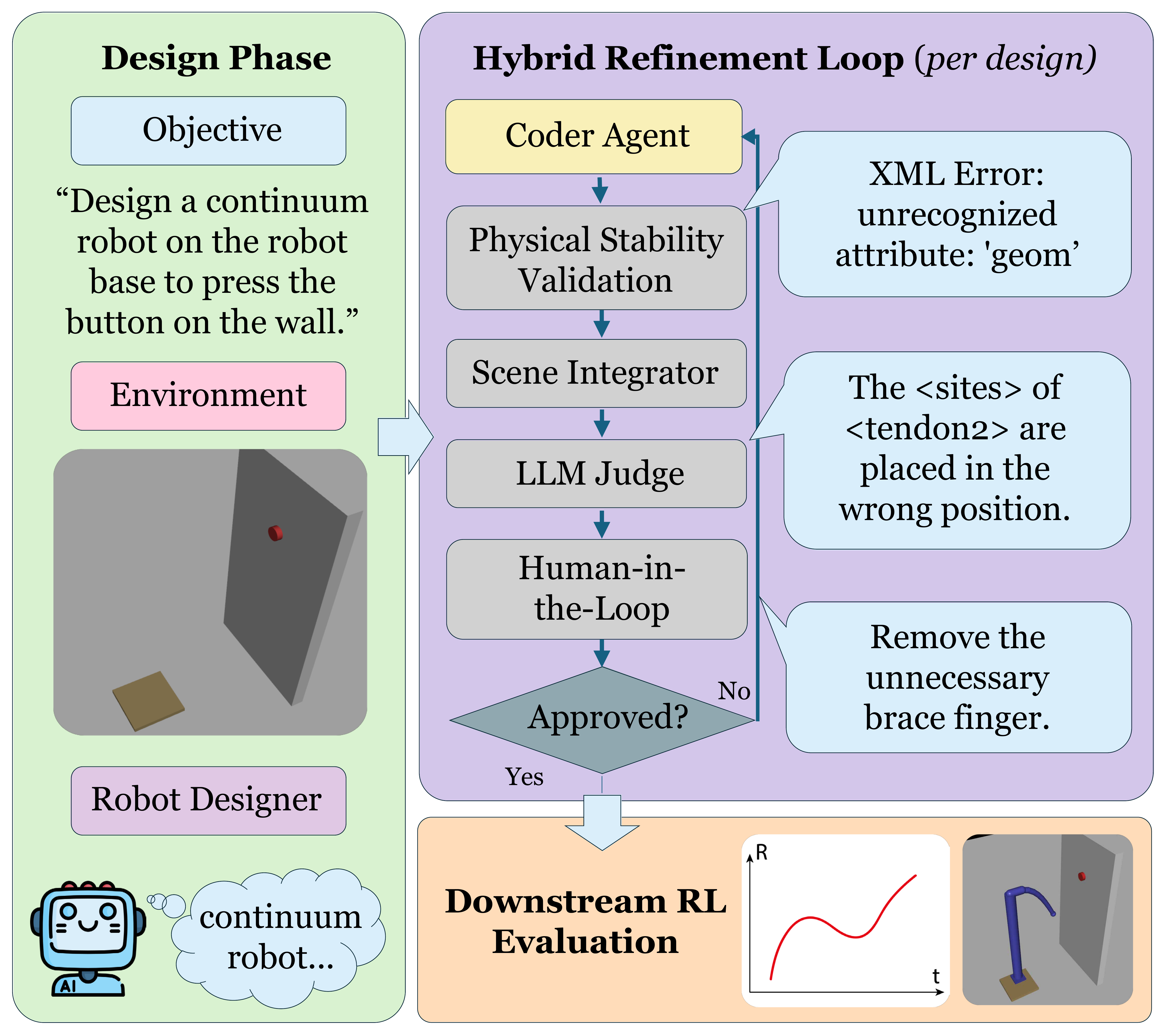}
    \caption{\textbf{The \framework\ Framework Overview.} The process begins with a user-defined objective. The \textbf{Designer} agent generates multiple initial design concepts. Each concept enters a parallel \textbf{Refinement Loop}, where the \textbf{Coder} agent generates a MuJoCo XML file. This design is subjected to four layers of feedback: (1) fast, deterministic \textbf{Physical Stability Validation}, (2) specific background scene integration for task functionality, (3) semantic, task-oriented critique from the \textbf{LLM Judge}, and (4) high-level guidance from a \textbf{Human-in-the-Loop}. The refined design is either approved or sent back to the Coder for another iteration. Approved designs are then evaluated in a downstream RL task.}
    \vspace*{-5mm}
    \label{fig:framework_overview}
\end{figure}

A fundamental limitation lies in the mismatch between how LLMs reason and how physical systems behave. LLMs operate primarily through linguistic patterns and abstractions, whereas the consequences of physical actions, such as instability, self-collision, or contact failure, are difficult to describe exhaustively in text\cite{wang2025large, cherian2024llmphy}. As a result, when a generated design fails in simulation, LLMs often lack the grounding needed to understand what went wrong and why. This leads to brittle designs with low physical validity and limited practical relevance, particularly in domains involving soft bodies, contact-rich interactions, and high-dimensional actuation.

In this work, we investigate how LLMs can be systematically guided to design physically feasible robots for meaningful tasks in 3-Dimensional environments. Rather than relying on one-shot generation, we propose a structured, iterative framework, \framework\ (Autonomous and Iterative Design of Soft Robots), that incrementally exposes the LLM to the physical implications of its own designs. 
The core idea is to establish a structured feedback pipeline that integrates simulator-observed physical states, semantic critique and human-in-the-loop feedback to guide the LLM designer through iterative robot design refinement.
By repeatedly refining robot morphology and physical parameters based on concrete feedback, the LLM is able to progressively improve the physical validity and functional potential of its designs.

We evaluate this framework in the context of tendon-driven continuum robots, a representative class of soft robots that combines high degree-of-freedom with complex contact behavior, while remaining suitable for simulation-based analysis\cite{russo2023continuum}. Tendon-driven continuum robots feature flexible backbones actuated by tensioned tendons, offering rich expressiveness but posing significant challenges for design automation\cite{rao2021model}. Their sensitivity to geometric parameters, material properties, and tendon routing makes them a compelling testbed for studying physically grounded robot design with LLMs.

To evaluate our approach, we construct a benchmark of 14 representative tasks spanning reaching, grasping, locomotion, and complex manipulation.
For each generated robot design that passes physical validation, we apply a basic reinforcement learning approach. Across the 14 tasks, an average of 26.7\% of the robots achieve the success criterion.
Finally, to validate real-world feasibility, we fabricate and deploy three LLM-designed robots in physical experiments, demonstrating the contribution to real robot design beyond purely simulated results.
The contributions of this paper are threefold:
\begin{itemize}
    \item We systematically formulate and analyze the problem of \textbf{physically grounded robot design with large language models}, identifying the fundamental mismatch between language-based reasoning and physical system behavior, and clarifying the key challenges arising from task-oriented interaction, object contact, and physical embodiment.
    \item We propose a \textbf{multi-layered, physically grounded framework} that incrementally exposes large language models to the physical consequences of their designs, enabling effective and iterative generation of physically valid robot designs. We integrate a lightweight \textbf{human-feedback component}, allowing domain experts to efficiently inject tacit engineering knowledge without hand-authoring rules or templates.
    \item We demonstrate the effectiveness of the proposed framework on the \textbf{automated design of tendon-driven continuum robots}, achieving efficient simulation-based design and successful sim-to-real transfer, thereby advancing the practical design of robots with real-world relevance.
\end{itemize}

\section{Related Work}

\subsection{Automated Robot Design}
Early studies optimize robot designs by tuning parameters of predefined structures or components~\cite{li2023modular}, typically using genetic algorithms~\cite{strgar2024evolution} or gradient-based methods~\cite{matthews2023efficient}. These approaches depend on human-specified design spaces, which can either lead to inefficiency when overly broad, or limit expressiveness when overly constrained. With the emergence of generative AI, diffusion-based models have been proposed to learn from robot design datasets and generate robots of similar types~\cite{chan2024creation, wang2023diffusebot}. However, their performance is closely tied to the quality and diversity of the training data.

More recently, several works leverage LLMs to directly generate code representations of robot mechanical structures. Some studies focus on idealized 2D simulation environments~\cite{chen2025large, song2025laser, ma2024exploration}, where robots are composed of simple blocks with different materials. Others extend this paradigm to 3D simulation~\cite{fang2025robomore, qiu2024robomorph}, but remain limited to relatively simple multi-legged crawling robots. Although some generated designs can be physically fabricated~\cite{ma2024exploration}, these approaches generally operate within restricted design spaces and narrowly defined tasks. A key limitation is that LLMs have limited ability to perceive and interpret the consequences of their designs within simulation environments, which constrains their applicability to complex models and tasks.

\subsection{Continuum Robots as a Design Challenge}

Continuum robots are robotic systems with compliant backbones capable of continuous deformation, offering high adaptability and safety in contact-rich environments \cite{russo2023continuum, burgner2015continuum}. These properties make them suitable for applications such as minimally invasive surgery and manipulation in confined or unstructured spaces \cite{kolachalama2020continuum}. Among various actuation paradigms, tendon-driven continuum robots are prominent due to their compact structure and ability to generate versatile bending and twisting motions through tendon tension control \cite{rao2021model}.

Many soft robots exhibit large deformations and nonlinear behaviors that challenge conventional modeling methods \cite{qin2024modeling}. For tendon-driven continuum robots, a common approach is to approximate the deformable body with discretized backbone segments connected by joints or elastic elements \cite{bajo2011kinematics}, achieving a practical trade-off between physical fidelity and computational efficiency suitable for simulation and learning-based control.

Despite the structural regularization, tendon-driven segmented-backbone continuum robots retain considerable design flexibility. Simple combinations of rigid segments and tendons have enabled diverse morphologies, including octopus-inspired manipulators \cite{wang2025spirobs} and DNA-helix–like structures \cite{hu2020bioinspired, blumenschein2018helical}.

From the perspective of automated and language-based design, such robots constitute a challenging yet well-structured testbed. Their design space is amenable to symbolic and parametric representation, while remaining sufficiently rich to capture the high-dimensionality and complex contact behaviors of soft robots.

\section{Method}
\label{sec:method}

In this section, we first formulate the problem of autonomous robot design and highlight the limitations of naive generative approaches~\ref{sec:preliminaries}. 
We then provide an overview of our proposed agentic framework, \framework, as illustrated in Figure~\ref{fig:framework_overview}. We detail its core design pipeline, which consists of an initial design conception stage and a multi-layer feedback loop enabling iterative refinement in~\ref{sec:framework}. 
Next, we further define the design space by specifying the structural and functional constraints of tendon-driven continuum robots, which constitute the target domain of our experiments in~\ref{sec:design_space}. 
Finally, we describe the evaluation protocol for assessing task completion, using reinforcement learning to verify convergence and task success beyond physical validity in~\ref{sec:downstream_eval}.

\subsection{Problem Formulation}
\label{sec:preliminaries}

We formally define the task of autonomous robot design as follows: given a natural language high-level objective $O$ (e.g., ``design a robot that can crawl through a narrow tunnel'') and an environment specification $E$ in MuJoCo XML format, the goal is to generate a set of valid and functionally promising MuJoCo XML scenes $\{S_1, S_2, \dots, S_k\}$. Each scene $S_i$ must have a physically stable continuum robot morphology $R_i$, capable of learning a control policy to address the objective $O$.

To motivate our structured approach, we first investigated a baseline where a state-of-the-art LLM (GPT-5) was directly prompted to generate a complete MuJoCo scene from an objective. 
We directly provided it with the objective description $O$ and the environment file $E$, and then required it to generate one complete scene $S_1$.
This naive method proved ineffective, achieving a success rate of nearly 0\% in producing syntactically correct and physically stable files. The predominant failure modes included generating invalid XML syntax, defining physically unstable structures that immediately collapse or explode in simulation, and creating morphologies functionally irrelevant to the task. This result highlights that successful generation requires a structured process of validation and iterative refinement, rather than simple, one-shot translation.

\subsection{The \framework\ Framework: An Agentic Workflow for Soft Robot Design}
\label{sec:framework}
We introduce our framework, \framework\ (Autonomous and Iterative Design of Soft Robots), depicted in Figure~\ref{fig:framework_overview}. \framework\ is founded on the principle of decomposing the complex design task into specialized roles handled by different agents within a closed-loop system. This agentic structure enables a robust and systematic exploration of the design space.

\textbf{Initial Design Conception.} The process begins with a \textit{Designer} agent that interprets the user's objective $O$ to propose several distinct, high-level design concepts. This initial step is crucial for promoting design diversity, preventing premature convergence to a single, potentially suboptimal idea. By seeding the process with a range of creative starting points (e.g., for the same locomotion task, they can be single-body peristaltic robots, multi-legged crawling robots, ring-shaped or other unconventional morphologies, etc.), the framework is able to explore disparate regions of the vast design space in parallel. After the designs are created, we utilize a \textit{Coder} agent to generate an initial version of the robots $\{R_1, R_2, \dots, R_k\}$ based on the original designs.

\textbf{The Hybrid Feedback and Refinement Loop.} This loop is the core of our framework, where each initial design is iteratively improved and validated. It integrates four critical feedback layers:

\underline{\textit{\ding{172} Physical Stability Validation:}} This component serves as the key bridge between language-driven design generation and physical feasibility. Each generated robot design $R_i$ is automatically and deterministically validated through direct interaction with the MuJoCo simulator, emulating the diagnostic workflow of a human robotics engineer. The design produced by the \textit{Coder} agent is first loaded into the simulator, where load-time failures such as syntax errors or invalid model definitions are immediately detected and returned to the Coder for correction. If instantiation succeeds, the model undergoes a targeted simulation phase in which predefined actuator tests are executed while sensor outputs and simulator states are monitored. Abnormal behaviors including constant sensor readings, high-frequency oscillations, or unexpected simulator resets, are analyzed using predefined diagnostic rules to identify common physical issues, such as ineffective actuation, self-collisions, or numerical instability. Based on these diagnostics, the design is classified as physically stable or unstable, and structured feedback is provided to guide iterative refinement. This automated process replaces manual debugging and enables large-scale, iterative robot design with LLMs.

\underline{\textit{\ding{173} Scene Integration:}} In this stage, the \textit{Scene Integrator} agent takes the generated (and correct) robot file $R_i$, the original environment file $E$ and task objective $O$ as input. Its role is to adapt the robot to the target environment and task by assessing their compatibility and adjusting integration-specific details, such as scaling and placement. The agent then merges the robot and environment descriptions into a unified scene file $S_i$ representing the robot operating within the environment. The resulting scene is subsequently subjected to the same physical stability validation procedure. If instability or simulation errors are detected, the integration process is repeated until a valid scene is obtained. This staged integration strategy decouples robot generation from environment-specific constraints, allowing the model to focus on coherent robot design before task and environment-level adaptation, which empirically leads to more reliable and stable outcomes.

\underline{\textit{\ding{174} Semantic Critique via LLM Judge:}} Designs that successfully pass the integration and physical validation stages are subsequently reviewed by a \textit{Judge} agent, which serves as the final quality gate before human involvement. The Judge assesses whether the complete design file is coherent, well-formed, and sufficiently meaningful to warrant downstream manual check or reinforcement learning, with the explicit goal of avoiding unproductive use of human effort. This evaluation follows a set of predefined semantic criteria, including consistency with the intended design space (e.g., adherence to the definition of a tendon-driven continuum robot), completeness of essential simulation components such as sensors and actuators, and basic feasibility with respect to the task objective and environment. Based on its assessment of the design file as a whole, the Judge assigns a pass/fail decision. Any designs that fail are accompanied by a structured critique which focuses on which components need improvement and are returned to the earlier agents for revision, while only qualified designs are forwarded for human inspection.

\underline{\textit{\ding{175} Human-in-the-Loop Refinement:}} The framework supports human-in-the-loop refinement after a complete and validated design $\{S_1, S_2, \dots, S_k\}$ has been produced. At this stage, a human engineer may inspect the generated design and provide high-level feedback in natural language, which is then routed back to the initial agent to guide subsequent regeneration. Importantly, this step is not necessary for the framework to function: the automated pipeline can terminate with a final output or continue iterating autonomously without human intervention. However, incorporating human feedback enables the injection of expert judgment and subjective preferences that are difficult to infer automatically. Experimental results demonstrate that such human involvement can improve design quality, highlighting a flexible trade-off between automation and expert labor.

Through iterative execution of the proposed framework, a set of valid and well-formed design files is generated. These qualified designs are subsequently forwarded to downstream reinforcement learning experiments for further evaluation. To ensure computational efficiency and robustness, an upper bound is imposed on the number of framework iterations. If a design repeatedly fails to pass the validation and evaluation stages within this limit, the generation process is terminated, and the corresponding attempt is recorded as an invalid design.

\begin{figure*}[hbt!]
    \centering
    \includegraphics[width=\linewidth]{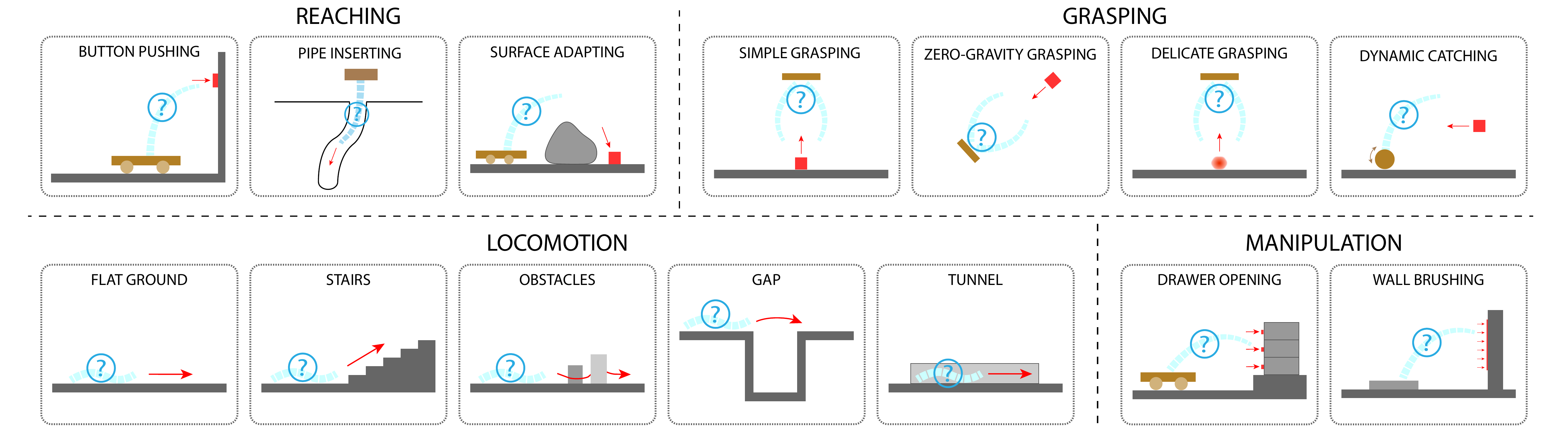}
    \caption{The test suite consists of 14 specific tasks across 4 categories. In the schematic illustration, the blue components denote the soft robot, whose detailed design is entirely generated by the LLM; the brown components represent a robot base that can move in predefined ways; the gray components indicate fixed structures in the environment; and the red elements denote the task objectives.}
    \vspace*{-5mm}
    \label{fig:all14_2D}
\end{figure*}

\subsection{Design Space Definition}
\label{sec:design_space}

In this work, we define the design space of tendon-driven continuum robot as follows: a soft structure that leverages under-actuation to achieve passive compliance, characterized by multiple body segments connected via ball or hinge joints and driven by one or more tendons to produce global shape deformation. After providing code examples to help the LLM understand this mechanical unit, we prompt it to autonomously select primitive geometries, assemble them into a robot, and determine suitable joint parameters, actuator layouts, actuator types, and corresponding control parameters. To ensure that the LLM generates syntactically valid XML files during iterative design, we also integrate the MuJoCo modeling specification into the prompt.

During the design generation stage, we provide only a brief task description as the objective $O$ for each scenario, without offering any further hints regarding mechanical configuration or implementation strategies. We instruct the LLM to first articulate its task interpretation and mechanical design rationale, and then produce a complete XML file through the iterative framework loop, which is subsequently fed into the simulation framework for verification and further refinement.

\subsection{Performance Evaluation}
\label{sec:downstream_eval}

To evaluate the functional quality of the robot morphologies generated by our framework, we adopt a standardized downstream evaluation pipeline based on reinforcement learning. 
The goal of the RL training is not to invest extensive human effort or optimization to achieve the best performance for each robot morphology. Instead, for all robot designs performing the same task, we employ basic and uniform RL training to demonstrate their potential to accomplish the task, thereby validating the effectiveness of the proposed framework.

\textbf{Reinforcement Learning.} Across all tasks, we employ the same reinforcement learning algorithm, Proximal Policy Optimization (PPO) \cite{schulman2017proximal}, to ensure consistency and comparability. While the learning algorithm is fixed, each task defines its own observation space, action space, reward function, and task-specific success criteria, reflecting the differing objectives and interaction requirements of reaching, grasping, locomotion and manipulation. Detailed specifications of these task formulations are provided in the appendix. For each auto-generated design, the corresponding task configuration is passed to a unified reinforcement learning training pipeline and trained using the same training algorithm, hyperparameters, and number of environment steps. After training, the learned policy is evaluated against predefined task success criteria.

\textbf{Reward Details.} Notably, we avoid morphology-specific reward shaping. Instead, all designs are trained under a \textit{single, generic reward formulation} that provides only sparse, task-level feedback based on progress toward task completion which reflects the practical constraint that the framework generates a large number of structurally diverse continuum robots, for which manual reward engineering would be infeasible. More importantly, it ensures a fair and consistent evaluation across designs, as no individual robot benefits from customized reward signals tailored to its specific morphology or dynamics. 

\textbf{Design Validity.} A design is also considered invalid if the robot stays still or produces implausible motion, or the simulator crashes during the training or evaluation process. Designs that complete training without triggering any such failure modes are further confirmed through manual inspection to check if they are structurally plausible continuum robots, and are classified as valid designs. Among valid designs, those that satisfy the task-specific success criteria during evaluation are classified as successful designs. This two-stage evaluation protocol allows us to distinguish failures due to infeasible morphology from those due to control difficulty, and provides a principled measure of the practical utility of LLM-generated robot designs.

\section{Experiments}
\label{sec:experiments}

\subsection{Task Design}

To evaluate the ability of LLMs to generate functional soft robotic designs, we construct a comprehensive benchmark consisting of diverse task scenarios and systematic experiments (Fig.~\ref{fig:all14_2D}). The benchmark is organized into four major task categories, inspired by prior studies on soft and continuum robots, and includes a total of fourteen subtasks. Each subtask is instantiated with multiple scenario variations to assess robustness and generalization. 

The first category of tasks is \textit{reaching}: controlling the robot’s deformation so that certain functional body parts reach specified target points within specific environments, typically within a workspace comparable to the robot’s own scale\cite{rao2021model}\cite{mao2024magnetic}. This task evaluates the LLM’s understanding of how a soft robot can achieve purposeful deformation.

The second category is \textit{grasping}: one of the most common tasks in robotics is using a gripper to properly pick up an object, maintain a stable hold, and deliver it to a target location\cite{lee2024fabrication}\cite{fishman2021dynamic}\cite{fu2025origami}\cite{zhang2023progress}\cite{wang2025spirobs}\cite{sinatra2019ultragentle}. This task evaluates the LLM’s understanding of how to design one of the most typical soft robot end-effectors. Note that here we do not consider the subsequent placement phase, though it is also a highly valuable direction for future exploration.

The third category is \textit{locomotion}: controlling the robot’s movement through specific environments, so that its entire body travels to a destination far beyond its own body scale\cite{li2023development}\cite{feng2025impulsive}. This task examines the LLM’s comprehension of how a soft robot can generate and sustain overall motion.

The fourth category involves \textit{complex manipulations}, where we design two relatively sophisticated tasks to explore the potential of LLMs. The first is drawer opening: the soft robot must repeatedly pull out and push in three drawers arranged vertically\cite{jiang2021hierarchical}. The second is force-controlled brushing: the soft robot must actively brush an entire wall surface while maintaining a controllable contact force\cite{chen2024vision}.

\subsection{Experimental Setup}
\label{sec:exp_setup}
All agents in our framework are powered by GPT-5, unless specified otherwise in our ablation studies. We set a maximum of 8 refinement iterations for each design concept and fully generate 3 concepts/designs per run. Each LLM call within the framework is limited to a maximum of $20000$ tokens, but each call usually consumed fewer than this limit. For each benchmark task, we generate 15 different designs to completion. 

\begin{figure*}[hbt]
  \centering
  \includegraphics[width=\textwidth]{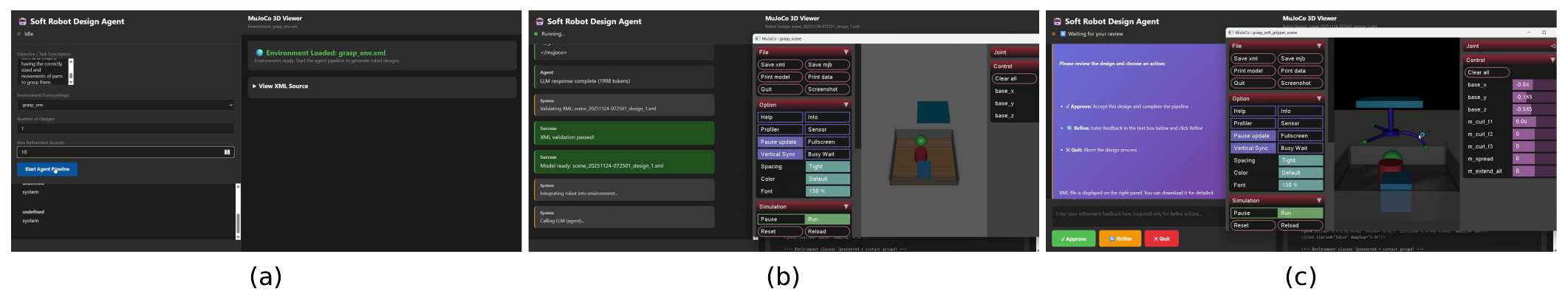}

  \vspace{-0.01ex}
  \caption{\textbf{Web-based interface for \framework.}
   The UI supports task/environment configuration, live monitoring of the agent pipeline with MuJoCo visualization, and human-in-the-loop decision making (approve/refine/quit). (a) Setup and launch. (b) Generation and simulation monitoring. (c) Review: approve or refine with feedback. With manual text input, each refinement only takes minutes.}
  \label{fig:aid-sr-ui}
  \vspace{-2.0ex}
\end{figure*}

To enable practical inspection and iterative steering, we implement a lightweight web interface that exposes the current MuJoCo scene, agent logs, and the generated XML artifact for each design iteration.
As shown in Fig.~\ref{fig:aid-sr-ui}, users can (a) configure the task/environment and launch the pipeline, (b) monitor the agent execution and simulation after XML validation, and (c) approve a design or provide text feedback to trigger another refinement round. An end-to-end screen recording is provided in the supplementary video.

The simulation experiments are conducted using the MuJoCo 3.3.4 physics engine\cite{todorov2012mujoco}, wrapped via the Gymnasium 1.1.1 interface\cite{gymnasium}. 
The observation space is defined as a hybrid feature vector that combines two types of information. One part consists of general proprioceptive data, including the robot’s base position, joint angles, and tendon-related measurements. The other part includes expert-designed, task-specific features, such as the relative distance to the target and readings from force sensors.
Additionally, the reward functions are customized for each specific task to guide the agent toward the desired behaviors. We employ the deep reinforcement learning algorithms provided by the Stable Baselines3 (SB3) library\cite{raffin2021stable} to address these tasks. Specifically, we utilize the Proximal Policy Optimization (PPO) algorithm\cite{schulman2017proximal}. 
After the LLM produces a batch of soft robot design proposals, we evaluate these generated designs through multiple testing methods.

For \textit{Reaching and Locomotion}, policies are learned directly through basic reinforcement learning via unconstrained exploration in the original action space. In the button-pushing and surface adapting tasks, we design the reward functions to encourage the robot to deform and bring the designated functional part closer to the target position. In the pipe inserting task, the policy is rewarded for moving the robot’s distal end downward while penalizing excessive contact forces against the pipe wall. For locomotion tasks, the rewards are primarily distance-based, guiding the robot toward minimizing its distance to the target position.

For \textit{Grasping}, we observe that many LLM-generated designs employ symmetric finger structures, where each finger is equipped with a dedicated actuator. When these actuators are controlled independently, exploration in the raw action space makes it difficult to efficiently discover coordinated grasping behaviors. To improve efficiency, we first examine the design rationale provided by the LLM to identify such symmetric structures. Based on this interpretation, actuators corresponding to symmetric fingers are grouped and controlled jointly, resulting in structured action definitions that better reflect the intended grasping mechanism of each design. 

\textit{Manipulation} involves two relatively complex operation sequences, for which we adopt sequential reward shaping. In the drawer opening task, we sequentially reward the robot for approaching the drawer, executing a grasp, and successfully pulling it open. In the wall brushing task, which requires force control, we reward the robot for following a predefined trajectory to brush the wall while maintaining contact forces within a desired range.

\textbf{Evaluation Metrics.} We use two primary metrics to evaluate our framework's performance:
(1) \textit{Valid Rate}, defined as the percentage of generated designs that are classified as valid after training and evaluation, i.e. designs that complete the simulation without failure and are structurally plausible continuum robots; (2) \textit{Success Rate}, defined as the percentage of generated designs that successfully accomplish the target task during evaluation (success criteria for different tasks are in the Appendix). We report three metrics to evaluate token efficiency: \textit{Average Token Count}, the total number of tokens (prompt + response) consumed to generate a robot design end-to-end; \textit{Average Cost} (\$), the corresponding cost given the token usage and the LLM pricing; and \textit{Average Latency} (s), the wall-clock time required to generate a robot design. For our main results (Table \ref{tab:main_results}) and other tables, we focused on the version of \framework\ without the human feedback component (unless otherwise stated) to keep our evaluation consistent and unbiased.

\begin{table*}[h!]
\centering
\caption{Main performance of LLM-designed robots across 14 benchmark tasks. These metrics reflect the performance of our framework without the human feedback component, and all token efficiency metrics are reported per design.}
\label{tab:main_results}
\begin{tabular}{llcccccc}
\toprule
\textbf{Category} & \textbf{Task Name} & \textbf{Valid Rate} & \textbf{Success Rate} & \textbf{Avg. Tokens} & \textbf{Avg. Cost(\$)} & \textbf{Avg. Latency(s)} \\
\midrule

\multirow{3}{*}{Reaching} & Button Pushing & 100\% & 78.67\% & 104137 & 0.57 & 472.38 \\
 
 & Surface Adapting & 100\% & 13.3\% & 123515 & 0.64 & 499.22 \\
 & Pipe Inserting & 93.3\% & 21.4\% & 212402 & 0.90 & 582.97 \\
\midrule
\multirow{4}{*}{Grasping} & Simple Grasping & 93.3\% & 33.3\% & 131211 & 0.65 & 477.11 \\
 & Zero-Gravity Grasping & 100\% & 13.3\% & 124512 & 0.62 & 444.12 \\
 & Delicate Grasping & 86.7\% & 20.0\% & 213250 & 0.93 & 602.97 \\
 & Dynamic Catching & 100\% & 26.7\% & 124512 & 0.62 & 484.75 \\
\midrule
\multirow{5}{*}{Locomotion} & Flat Ground  & 100\% & 46.7\% & 103321 & 0.54 & 410.92 \\
 & Stairs & 100\% & 26.7\% & 119281 & 0.60 & 443.40 \\
 & Obstacles & 100\% & 13.3\% & 143186 & 0.72 & 510.39 \\
 & Gap & 93.3\% & 53.3\% & 182576 & 0.81 & 530.75 \\
 & Tunnel & 100\% & 33.3\% & 185429 & 0.88 & 616.59 \\
\midrule
\multirow{2}{*}{Manipulation} & Drawer Opening & 100\% & 6.7\% & 90756 & 0.48 & 385.51 \\
 & Force-Controlled Brushing & 80\% & 0\% & 261151 & 1.19 & 867.89 \\
\bottomrule
\end{tabular}
\end{table*}

\subsection{Framework Performance on Benchmark Tasks}
\label{sec:main_results}

Based on the experimental results in Table \ref{tab:main_results}, the proposed framework can efficiently generate syntactically valid design files; however, due to the inherent limitations of current LLMs, only a small fraction of the generated designs are able to successfully complete the tasks.

Among all tasks, the \textit{Reaching} category exhibits the highest success rate. This is primarily because these tasks require relatively simple structures: starting from the given robot base, it is generally sufficient to construct a mechanism with adequate workspace and a reasonable range of motion. The \textit{Locomotion} tasks show the next-highest success rate, likely for two reasons. First, these tasks require coherent global motion of the entire structure, imposing higher demands on the overall plausibility of the mechanical design. Second, since a scalable morphology-aware reward scheme for reinforcement learning is not currently feasible, training is unable to fully exploit the potential performance of each generated design. The success rate of the \textit{Grasping} tasks is further reduced. Due to the LLM’s limited understanding of global morphology and physical contact, it rarely produces complete, well-fitted, and structurally stable soft grippers. Finally, the results of the two complex \textit{manipulation} tasks highlight the limitations of current LLMs even more clearly. When faced with sophisticated task sequences and the corresponding mechanical requirements, the model lacks sufficient understanding and generation capability, and therefore produces practically viable designs only with very low probability.

Table~\ref{tab:main_results} also reports token-efficiency metrics per design, capturing the end-to-end LLM interaction required to produce a valid robot candidate. Across all 14 tasks, our framework uses on average 151K tokens per design, corresponding to an average cost of (\$0.73) per design and 523s latency. Most tasks cluster in a practical band of roughly 100–185K tokens ((\$0.54)–(\$0.88), 411–617s), indicating predictable overhead across reaching, grasping, and locomotion, while more interaction-heavy settings can increase budget (e.g., Force-Controlled Brushing reaches 261K tokens ((\$1.19), 868s)) which corresponds with the framework needing to undergo many more iterations to adequately pass all of the refinement checks. Overall, these results suggest that token usage (and thus cost/latency) scales with task difficulty and constraint complexity, while remaining within a consistent per-design budget for the majority of benchmarks.

\subsection{Ablation Studies and Component Analysis}
\label{sec:ablations}

We perform a series of targeted ablation studies to analyze the proposed framework and quantify the contributions of its key components.

\textbf{Impact of the Hybrid Feedback System.}
To isolate the effect of each feedback layer, we compare the \textit{Full Framework} against five ablated variants on a subset of representative tasks. These variants include: (1) \textit{Human Feedback}, which performs a single refinement round using only human feedback; (2) \textit{No Judge}, where iterative refinement relies solely on physical validation, without semantic critique; (3) \textit{No Scene Integrator}, in which the environment-aware scene integration component is removed; (4) \textit{No Judge or Scene Integrator}, which disables both components; and (5) \textit{No Validation}, where the Physical Stability Validation module is omitted, allowing designs to bypass MuJoCo-based feasibility checks. As reported in Table~\ref{tab:ablation_feedback}, the full framework achieves the highest overall performance. Removing the Judge results in a 10.7\%–20\% reduction in task success compared to the full framework without human feedback, indicating that the Coder agent struggles to produce meaningful functional improvements in the absence of structured semantic guidance. Similarly, the \textit{No Scene Integrator} and \textit{No Judge or Scene Integrator} variants exhibit degraded performance, underscoring the importance of environment-aware evaluation and component composition. The \textit{No Human Feedback} setting also leads to a noticeable performance drop, suggesting that human intuition remains beneficial for escaping local optima and aligning designs with nuanced task objectives. Finally, the \textit{No Validation} variant yields a 0\% physical validity rate, as none of the generated designs can be reliably instantiated or simulated in MuJoCo, highlighting the indispensable role of physical stability validation in generating viable robot designs.

\begin{table}[h!]
\caption{Ablation study on the hybrid feedback system for the \textit{Flat Ground} and \textit{Button Pushing} tasks. Note that this experiment is conducted separately from the overall experiments reported in Table~\ref{tab:main_results}.
We regenerate a batch of designs for each configuration, except that \textit{Full Framework (w/ human feedback)} is incorporated by adding human evaluations as textual inputs on top of the \textit{Full Framework (w/o human feedback)}, followed by one additional iteration of generation.}
\label{tab:ablation_feedback}
\begin{tabularx}{\columnwidth}{lcc}
\toprule
\textbf{Configuration} & \textbf{Valid Rate} & \textbf{Success Rate} \\
\midrule
\multicolumn{3}{c}{\textit{Button Pushing Task}} \\
Full Framework (w/o human feedback) & 100\% & 70.7\% \\
Full Framework (w/ human feedback) & 100\% & 76.0\% \\
No Judge & 100\% & 60.00\% \\
No Scene Integrator & 86.7\% & 53.3\% \\
No Judge or Scene Integrator & 93.3\% & 60.00\% \\
No Validation & 0\% & \textbackslash \\
\midrule
\multicolumn{3}{c}{\textit{Flat Ground Task}} \\
Full Framework (w/o human feedback) & 100\% & 53.3\% \\
Full Framework (w/ human feedback) & 100\% & 60\% \\
No Judge & 80\% & 33.3\% \\
No Scene Integrator & 80\% & 40\% \\
No Judge or Scene Integrator & 73.3\% & 33.3\% \\
No Validation & 0\% & \textbackslash \\
\bottomrule
\end{tabularx}
\end{table}

\textbf{Analysis of Different Backbone LLMs.}
We first evaluate the proposed framework using three state-of-the-art backbone LLMs: GPT-5, Gemini 2.5 Pro, and Claude Sonnet 4.5. All three models are able to operate effectively within the framework, with GPT-5 achieving the highest physical validity (100\%) and the best overall task performance, and is therefore selected as our default backbone. In our analysis, we prioritize physical validity as the primary comparison metric, as it more directly reflects the model’s ability to identify and repair errors under iterative physical and semantic constraints, whereas task success exhibits higher variance due to downstream training stochasticity. We also evaluate several recent open-source models and observe a clear degradation in physical validity for smaller-capacity backbones, which frequently fail to recover from invalid designs and remain trapped in repeated infeasible generations. This suggests that sufficient model capacity is important for satisfying the dense constraints and multi-step correction required by iterative robot design.

\begin{table}[!t]
\scriptsize
\centering
\caption{Performance comparison of different models on the \textit{Flat Ground} task. Valid rate better reflects reasoning and error-correction ability than success rate, while average tokens correlate with both reasoning depth and iterative failures; overall, larger models achieve superior performance.}
\label{tab:ablation_models}
\begin{tabularx}{\columnwidth}{lccc}
\toprule
\textbf{Model} & \textbf{Valid Rate} & \textbf{Success Rate} & \textbf{Avg. Tokens (per design)} \\
\midrule
GPT-5 & 100\% & 46.7\% & 103321 \\
Gemini 2.5 Pro & 73.3\% & 0\% & 113625 \\
Claude Sonnet 4.5 & 66.7\% & 26.7\% & 180761 \\
GPT-OSS-120B & 40\% & 0\% & 176755 \\
Llama3.3-70B & 33.3\% & 6.67\% & 117417 \\
Qwen3-30B & 0\% & \textbackslash & 168070 \\
\bottomrule
\end{tabularx}
\end{table}

\begin{figure}[!t]
    \centering
    \includegraphics[width=\linewidth]{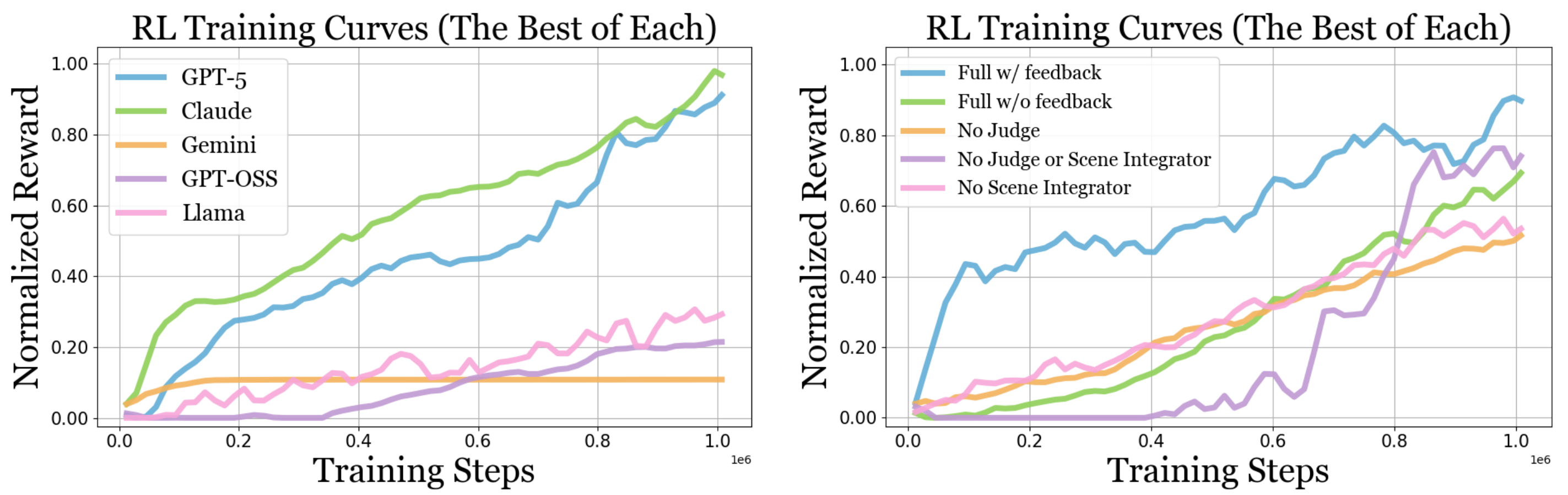}
    \caption{\textbf{Reward curves of the best-performing robots generated by different LLMs or frameworks.} Due to high uncertainty and randomness in individual robot design, differences in reward values do not directly indicate the performance of the LLMs or frameworks themselves; this figure mainly illustrates the effectiveness and stability of the reinforcement learning methods.}
\end{figure}

\subsection{Sim2Real Demonstration and Discussion}

In this work, we define the exploration space of the continuum robot and the corresponding task scenarios with physical feasibility in mind, aiming to encourage the LLM to generate designs that are not only functional in simulation but also potentially applicable in real-world settings. To demonstrate this feasibility, we select three high-performing designs discovered during simulation and reconstruct them according to their design files. These designs are then physically fabricated using 3D printing and assembled into real robotic systems, which are subsequently evaluated on the same tasks as in simulation.

\begin{figure}[!t]
    \centering
    \includegraphics[width=0.97\linewidth]{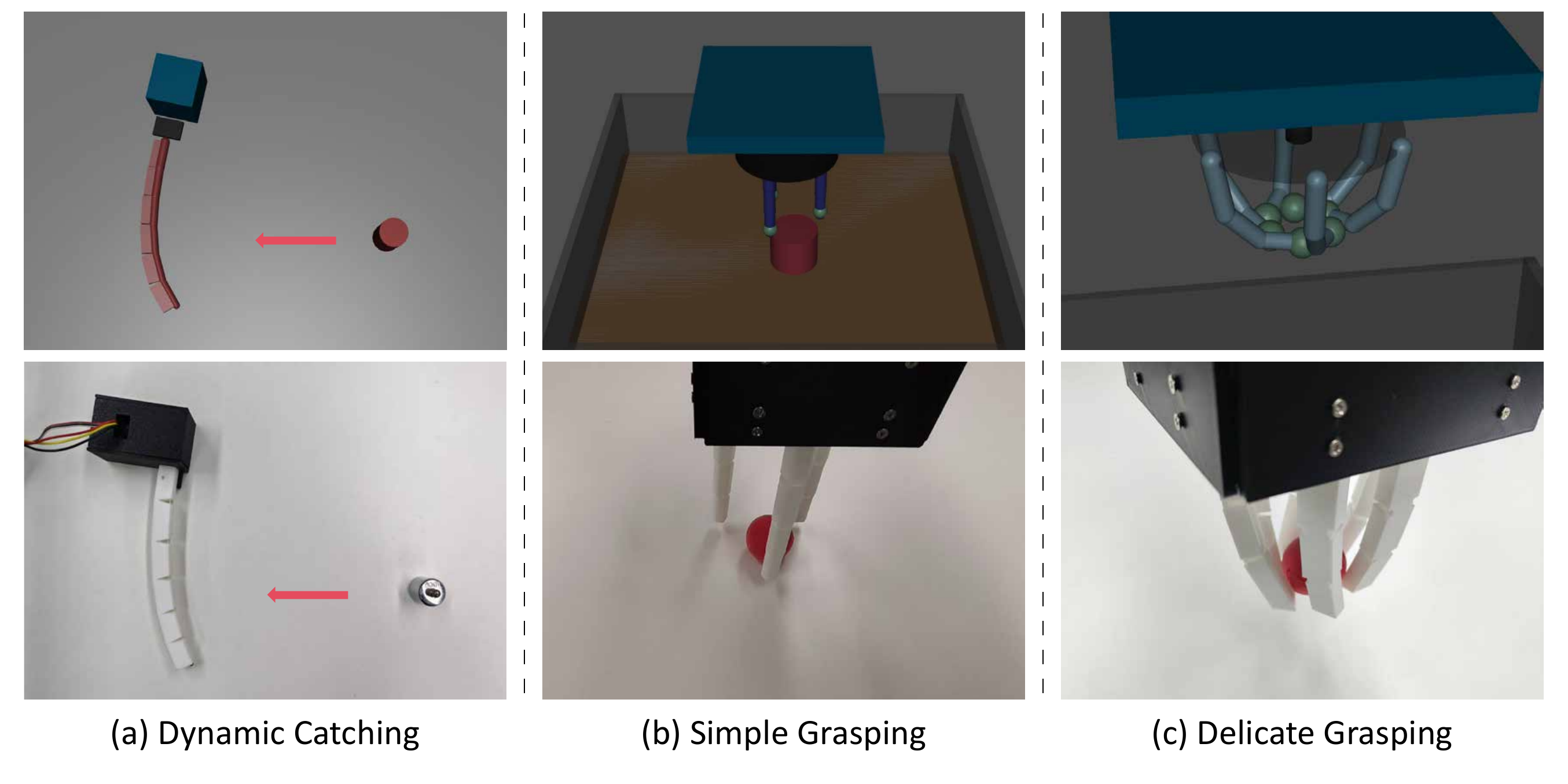}
    \caption{\textbf{Sim-to-real demonstration of three representative designs.}
\textbf{(a)} For dynamic catching, the LLM designed a continuum robot segmented into arithmetic lengths. This causes the robot to form a logarithmic spiral, exhibiting grasping capability.
\textbf{(b)} For simple grasping, the LLM designed a typical gripper composed of three continuum robots as fingers, each following the same design principle as in (a).
\textbf{(c)} For delicate grasping, the LLM proposed an intriguing structure consisting of six fingers, where each joint of every finger bends laterally, forming a helical shape. When all six fingers bend simultaneously, it generates a large contact area, enabling the enclosure of small objects with diverse shapes.}
\vspace*{-5mm}
    \label{fig:sim2real}
\end{figure}

Since the environment in these three tasks all have large robot base structures, they can accommodate motors as actuators; therefore, the fabrication is relatively straightforward. Qualitative comparisons between simulation and real-world execution are illustrated in Fig.\ref{fig:sim2real}, where the robots perform identical tasks in both domains. Despite inevitable modeling inaccuracies and unmodeled physical effects, the overall motion patterns, deformation behaviors, and task execution strategies exhibit a high degree of visual and functional consistency. 

These observations suggest that the design space defined for the language model possesses a meaningful degree of physical practicality. The ranges of key parameters, such as size, weight, and structural stiffness, largely fall within regimes that can be realized using common fabrication methods and commercially available components. As a result, the generated designs are not merely abstract artifacts of simulation but represent plausible robotic systems that can be instantiated and deployed. This sim-to-real consistency indicates that large language models, when guided by appropriately constrained design representations, can serve as effective tools for proposing physically grounded robot designs with potential practical value. 

However, several limitations remain apparent when trying to transfer designs from simulation to the real world. A number of designs that perform well in simulation cannot be directly realized in hardware due to idealized modeling assumptions. For instance, tendons in simulation are allowed to ignore intermediate body parts and directly connect distant attachment sites, whereas real cables or ropes must obey geometric constraints and physical routing. Similarly, simulated actuators can assume large gear ratios without accounting for actuator volume, mass, or thermal limits, resulting in power densities that exceed those of nearly all practical actuator technologies. These discrepancies highlight the gap between simulated expressiveness and physical realizability, and point to the need for more restrictive modeling of actuation and transmission mechanisms in future work. 

Overall, while a significant sim-to-real gap still exists, our framework succeeds in translating LLM generated designs into fabricated robots that perform meaningful real-world tasks. The successful deployment of three representative designs demonstrates that the LLM can generate robot morphologies with genuine physical competence, rather than designs confined to idealized simulation environments. These results indicate a concrete step toward bridging language-based reasoning and physical embodiment.

\section{Conclusion}
We explore a practical pathway for grounding large language models in physical robot design through a multi-layered, feedback-driven framework. By iteratively incorporating constraints, simulation feedback, and semantic critique, the LLM is able to better interpret the physical consequences of its own designs, resulting in consistently high physical validity. This suggests that such layered interaction is a promising and reusable strategy for LLM-based embodied design.
Applied to tendon-driven continuum robots, the framework produces several designs that successfully transfer from simulation to real-world operation, indicating its potential value as a design aid for more complex robotic systems.

Nevertheless, most of the generated robots tend to be relatively simple, reflecting both the limitations of current simulators and the LLM’s still-developing ability to propose unconventional yet reasonable designs. Additionally, the proposed framework can only provide guidance on valid designs, while it lacks reasonable feedback mechanisms regarding the task-level performance.
Addressing these limitations remains an important direction for future work.

\bibliographystyle{IEEEtran}
\bibliography{reference}

\end{document}